\documentclass[10pt,twocolumn,letterpaper]{article}

\usepackage[final]{cvpr}              

\usepackage[dvipsnames]{xcolor}

\definecolor{cvprblue}{rgb}{0.21,0.49,0.74}
\usepackage[pagebackref,breaklinks,colorlinks,citecolor=cvprblue]{hyperref}
\usepackage{graphicx}
\usepackage{amsmath}
\usepackage{amssymb}
\usepackage{booktabs}
\usepackage{adjustbox}
\usepackage{bbding}
\usepackage{makecell}
\usepackage{multirow}
\usepackage[accsupp]{axessibility}

\def\paperID{11898} 
\def\confName{CVPR}
\def\confYear{2024}

\title{DocRes: A Generalist Model Toward Unifying Document Image \\ Restoration Tasks}

\author{Jiaxin Zhang$^{1,2}$, Dezhi Peng$^{1}$, Chongyu Liu$^{1}$, Peirong Zhang$^{1}$, Lianwen Jin$^{1,2,}$ \thanks{Corresponding author}\\
{\normalsize $^1$ South China University of Technology}\\
{\normalsize $^2$ INTSIG-SCUT Joint Lab on Document Analysis and Recognition}\\}

\begin{document}
\maketitle
\begin{abstract}

Document image restoration is a crucial aspect of Document AI systems, as the quality of document images significantly influences the overall performance. Prevailing methods address distinct restoration tasks independently, leading to intricate systems and the incapability to harness the potential synergies of multi-task learning. To overcome this challenge, we propose DocRes, a generalist model that unifies five document image restoration tasks including dewarping, deshadowing, appearance enhancement, deblurring, and binarization. To instruct DocRes to perform various restoration tasks, we propose a novel visual prompt approach called \textbf{D}ynamic \textbf{T}ask-\textbf{S}pecific \textbf{Prompt} (DTSPrompt). The DTSPrompt for different tasks comprises distinct prior features, which are additional characteristics extracted from the input image. Beyond its role as a cue for task-specific execution, DTSPrompt can also serve as supplementary information to enhance the model's performance. Moreover, DTSPrompt is more flexible than prior visual prompt approaches as it can be seamlessly applied and adapted to inputs with high and variable resolutions. Experimental results demonstrate that DocRes achieves competitive or superior performance compared to existing state-of-the-art task-specific models. This underscores the potential of DocRes across a broader spectrum of document image restoration tasks. The source code is publicly available at \url{https://github.com/ZZZHANG-jx/DocRes}.

\end{abstract}

\section{Introduction}
\label{sec:intro}

\begin{figure}[t]
    \includegraphics[width=2.4in]{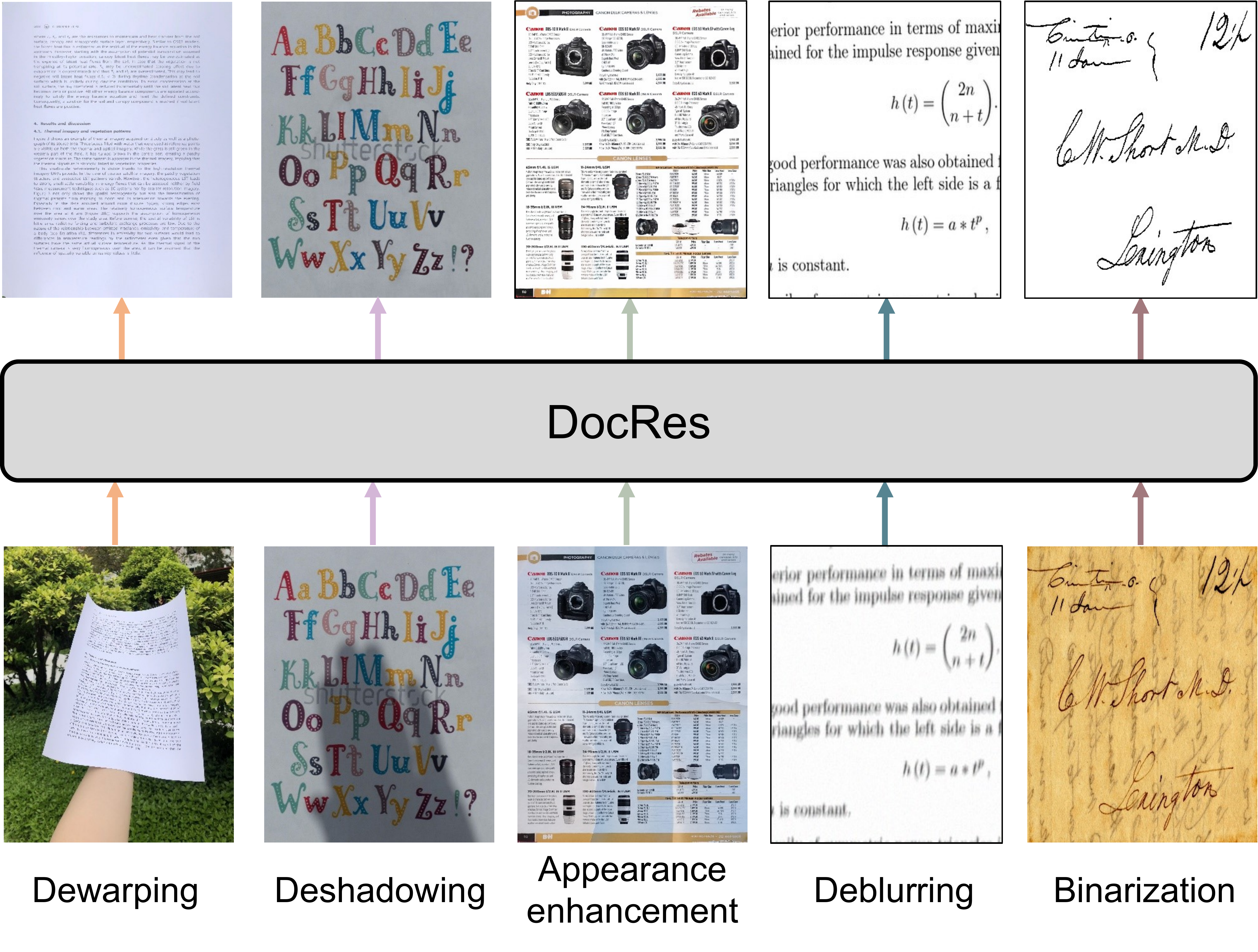}
    \centering
    \caption{DocRes is a generalist model that unifies five document image restoration tasks, including tasks of dewarping, deshadowing, appearance enhancement, deblurring, and binarization.}
    \label{fig:motivation}
\end{figure}

Photographed or scanned document images frequently manifest various forms of degradation, encompassing geometric distortion, shadows, bleed-through, stains, and more. These degradations pose substantial challenges for existing document analysis and recognition systems, and also significantly compromise the visual appeal and legibility of documents, as underscored in recent studies~\cite{feng2021doctr,wang2022udoc,zhang2023appearance,das2019dewarpnet,li2023foreground}.

The restoration of document images, addressing objectives like flattening documents, shadow removal, clean appearance restoration, deblurring, or text segmentation, holds both academic and practical significance. Existing approaches~\cite{feng2021doctr,das2019dewarpnet,wang2022udoc,ma2018docunet,yang2024gdb} typically treat these restoration tasks separately, relying on models specifically designed for each task. While effective in achieving commendable performance, this paradigm results in a system that requires the design of multiple models and extensive maintenance. Moreover, it fails to leverage the potential synergies of multi-task learning. 

To tackle this problem, motivated by recent pioneering works~\cite{wang2022ofa,wang2023images,chen2022unified,li2020all,lu2022unified,peng2023upocr} that unify various vision tasks, we seek to explore a generalist model for document image restoration. As illustrated in Fig.~\ref{fig:motivation}, our proposed DocRes seeks to unify five document image restoration tasks: dewarping, deshadowing, appearance enhancement, deblurring, and binarization. To empower such a generalist model to perform specific tasks and generate desired outputs, it is often crucial to convey instructional information to the model. Existing visual generalist models accomplish this by drawing inspiration from advancements in Natural Language Processing (NLP) and leveraging prompt learning techniques. Notably, studies like~\cite{chen2022unified, wang2022ofa, tang2023unifying} transform the vision task into an NLP one by discretizing continuous vision output and using discrete tokens as prompts. However, the autoregressive decoding paradigm they employ is inherently less effective for low-level tasks.

In the realm of unifying image-to-image tasks, recent methods~\cite{wang2023images, bar2022visual, liu2023unifying, ma2023prores} have suggested vision-centric prompts, known as visual prompts, which exhibit promising potential across various vision tasks, including image restoration. Among them, approaches like~\cite{wang2023images, bar2022visual, liu2023unifying} propose using a pair of input/output samples as a visual prompt to guide the model to perform the corresponding task. Nevertheless, these methods exhibit reduced efficiency as they necessitate an additional pair of samples during inference, making them less suitable for low-level tasks involving high-resolution images. ProRes~\cite{ma2023prores} introduces a more straightforward visual prompt method, wherein a matrix composed of learnable parameters, matching the input's shape, is assigned for each task and added to the input as a visual prompt. However, the training process of ProRes is intricate, which requires initializing each visual prompt by pre-training on task-specific models. Additionally, some of the above visual prompt methods rely on Mask Image Modeling (MIM) and require the ViT~\cite{dosovitskiy2020image} framework. However, ViT typically demands that input images during testing and training have the same resolution, and due to memory constraints, ViT struggles with high-resolution images. 
This makes it challenging for these methods to adapt to low-level tasks, which typically involve patch training, whole-image testing, as well as variable and high-resolution inputs.

Recognizing the limitations of the aforementioned prompt methods, we introduce a new visual prompt for DocRes called the \textbf{D}ynamic \textbf{T}ask-\textbf{S}pecific \textbf{Prompt} (DTSPrompt). The inspiration behind DTSPrompt is rooted in prevalent practices observed in existing document image restoration endeavors, where certain prior features extracted from the input image are typically employed to enhance model's performance, such as background images for shadow removal~\cite{zhang2023document,lin2020bedsr} and text content masks for dewarping~\cite{zhang2022marior,jiang2022revisiting,feng2022geometric}. Specifically, the DTSPrompt for different tasks comprises distinct prior features based on the characteristics of each task, where prior features we adopted include document segmentation masks, binarization results, gradient maps, and so on. DTSPrompt can not only serve as an effective cue for the model to determine which task to perform but also function as supplementary information to enhance the performance for the corresponding task. DTSPrompt can be seamlessly applied to various existing restoration networks, rather than limited to ViT. This enables DocRes to handle high and variable resolutions encountered in document restoration.

Experiments demonstrate that, without additional complex network designs, DocRes, as a generalist model, achieves competitive or even superior performance on various benchmarks compared to existing well-established and carefully designed task-specific methods. 

In summary, our contributions are as follows:

\begin{itemize}
  \item To the best of our knowledge, we are the first to explore generalist models for unifying restoration of document images, which serves as a pioneering effort in this field.
  \item We propose a simple yet highly effective visual prompt approach termed DTSPrompt, which extracts different prior features from the input image to create prompts. It can guide the model to distinguish different tasks, provide supplementary information to improve performance and accommodate both high and variable resolutions.
  \item Empowered by DTSPrompt, our DocRes achieves competitive or even better performance compared to task-specific methods across various benchmarks. 
\end{itemize}

\section{Related works}

\subsection{Document Image Restoration}

Dewarping~\cite{ma2018docunet,zhang2022marior,das2019dewarpnet,feng2022geometric,feng2021docscanner} addresses the elimination of geometric distortions such as curves and crumples, which not only hinder OCR engine performance~\cite{cheng2023m6doc,zhang2023appearance,wang2023scene} but also degrade document readability. Deshadowing~\cite{lin2020bedsr,zhang2023document,li2023high} focuses on removing shadows, a common occurrence in photographed document images, to produce shadow-free documents. Appearance enhancement, also known as illumination correction~\cite{zhang2023appearance,wang2022udoc,das2020intrinsic}, goes beyond specific appearance degradations, striving to restore a clean appearance akin to digital-born PDFs. This is valuable as it significantly enhances document readability and aesthetics. Deblurring~\cite{hradivs2015convolutional,souibgui2020gan,yang2023docdiff} aims to eliminate blurriness and restore a clear image. Binarization~\cite{yang2024gdb} involves segmenting foreground text from document images, a critical task for applications primarily focused on text content, usually obscured by stains, artifacts, black margins, or weak contrast.

Existing approaches~\cite{zhang2022marior,das2019dewarpnet,zhang2023document,lin2020bedsr,zhang2023appearance,wang2022udoc,yang2024gdb} typically treat these restoration tasks independently, resulting in a complex document image restoration system and the inability to harness the potential synergies among tasks. While recent efforts~\cite{yang2023docdiff,souibgui2020gan,souibgui2023text} aim to tackle several tasks with a unified network architecture, they still require individual training and separate models for each task. 

\subsection{All-in-one image restoration}
Unlike the domain of document image restoration that lacks unification, recent studies~\cite{li2020all,valanarasu2022transweather,ma2023prores,wang2023images,liu2023unifying,zhang2023all} have made pioneering strides in developing generalist models for natural scene image restoration. They encompass tasks such as weather effect removal, low-light image enhancement, denoising, and deblurring. Existing multi-task generalist models can be broadly classified into two categories based on whether the task is explicitly specified for the model during inference: task-agnostic and task-oriented. 

\textbf{Task-agnostic}. Task-agnostic methods~\cite{kulkarni2022unified,li2022all,zhang2023all,valanarasu2022transweather,potlapalli2023promptir} do not require users to specify the task type, but they are less flexible and cannot handle a broader range of tasks, typically being confined to specific domains like weather effect removal. This arises because some tasks share similar inputs but demand distinct outputs, leading to ambiguity in the learning process. The task-agnostic setup proves inappropriate for unifying document image restoration since tasks like dewarping, deshadowing, and appearance enhancement share similar inputs but demand distinct outputs.

\textbf{Task-oriented}. To achieve a task-oriented generalist model, explicit task information needs to be introduced. Some approaches~\cite{wang2022ofa,wang2023images,chen2022unified,li2020all,lu2022unified} discretize the continuous vision output, use discrete tokens as task prompts, and use the Transformer decoder for autoregressive prediction. These methods are more suited for visual understanding tasks, such as detection, captioning, and visual question answering but are inappropriate for low-level tasks involving high-resolution outputs. More recently, there have been methods proposing vision-centric prompts~\cite{wang2023images,ma2023prores,liu2023unifying,bar2022visual} for unification purposes, known as visual prompts, showing promising potential in various vision tasks, including image restoration. Among them,~\cite{wang2023images,liu2023unifying,bar2022visual} use an input/output sample of a specific task as a prompt, resembling the paradigm used in inpainting tasks where surrounding pixel information is learned to fill in missing pixel positions. Nevertheless, the additional input leads to inefficiency and limitation of image resolution. ProRes~\cite{ma2023prores} employs learnable parameters with the same shape of the input image as a visual prompt and adds it pixel-wise to the input. While effectively guiding the model, the prompt for each task needs to be trained on task-specific models for initialization, resulting in a complex training pipeline. 

Moreover, all these visual prompt methods mentioned above are limited to the ViT~\cite{dosovitskiy2020image} framework, making them unable to adapt to the variable resolutions in restoration tasks. The quadratic computational complexity also restricts the input resolution. For example, the input resolution for ProRes~\cite{ma2023prores} and Painter~\cite{wang2023images} is limited to $448\times448$, which is insufficient for document images with resolutions commonly exceeding 1K. While block-wise processing could be employed, many document image restoration tasks heavily depend on global information, such as deshadowing and binarization, making this strategy unfeasible. In contrast, our DTSPrompt method is not confined to the ViT framework and can be applied to various more flexible restoration networks to form our DocRes model.

\section{Methodology}
\begin{figure*}[t]
    \includegraphics[width=5.3in]{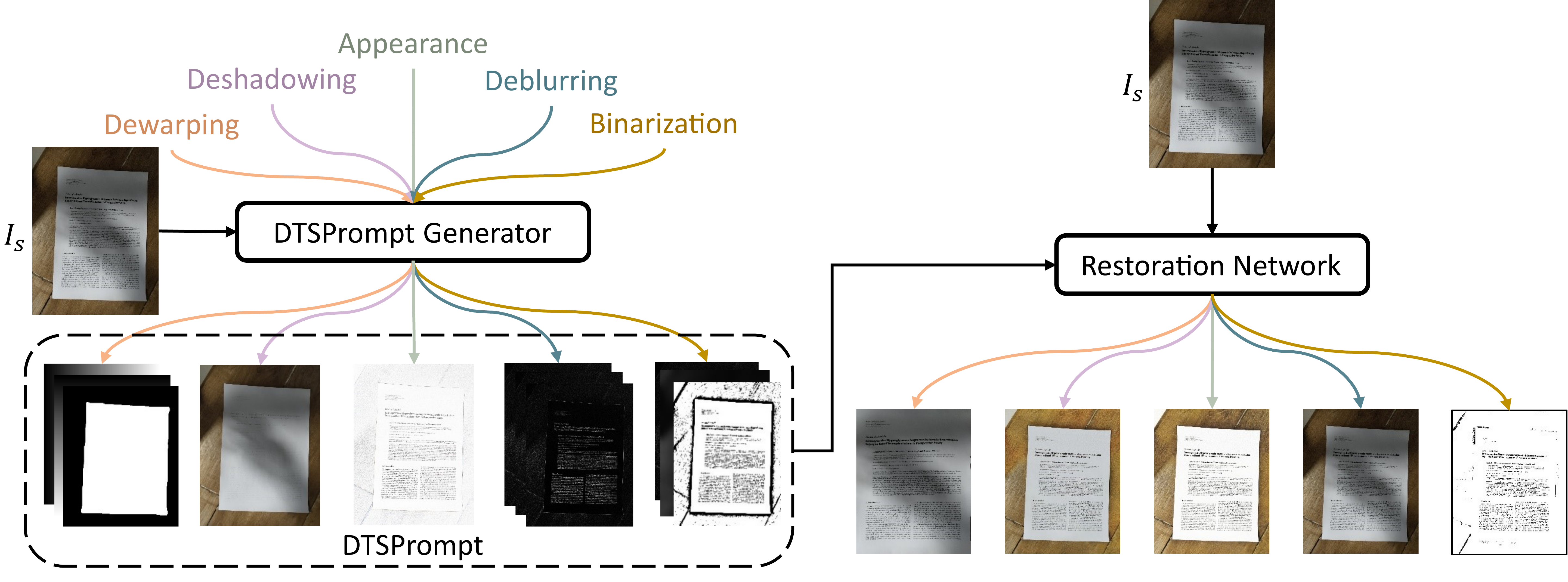}
    \centering
    \caption{The overall pipeline for DocRes. The document image to be restored, denoted as $I_s$, is initially fed into the DTSPrompt generator, which extracts specific prior features based on the task to form the DTSPrompt. Alongside $I_s$, DTSPrompt is input into the restoration network. It serves not only as a guidance for the restoration network on the particular task to be performed but also functions as auxiliary information derived from $I_s$ to improve performance.}
    \label{fig:architecture}
\end{figure*}

As depicted in Fig.~\ref{fig:architecture}, the input document image undergoes an initial processing step by the DTSPrompt generator to generate task-specific DTSPrompts, which are composed of various prior features extracted from the input image. Such DTSPrompts are instrumental in guiding the restoration network to execute distinct tasks while simultaneously enhancing overall performance. In the following sections, we first provide a detailed exploration of the process involved in obtaining various prior features to construct the DTSPrompt for each task. Then we introduce our prompt fusion approach and the selection of the restoration network.

\subsection{Dynamic task-specific prompt}
\label{sec:prompt}

\begin{figure}[t]
    \includegraphics[width=2.7in]{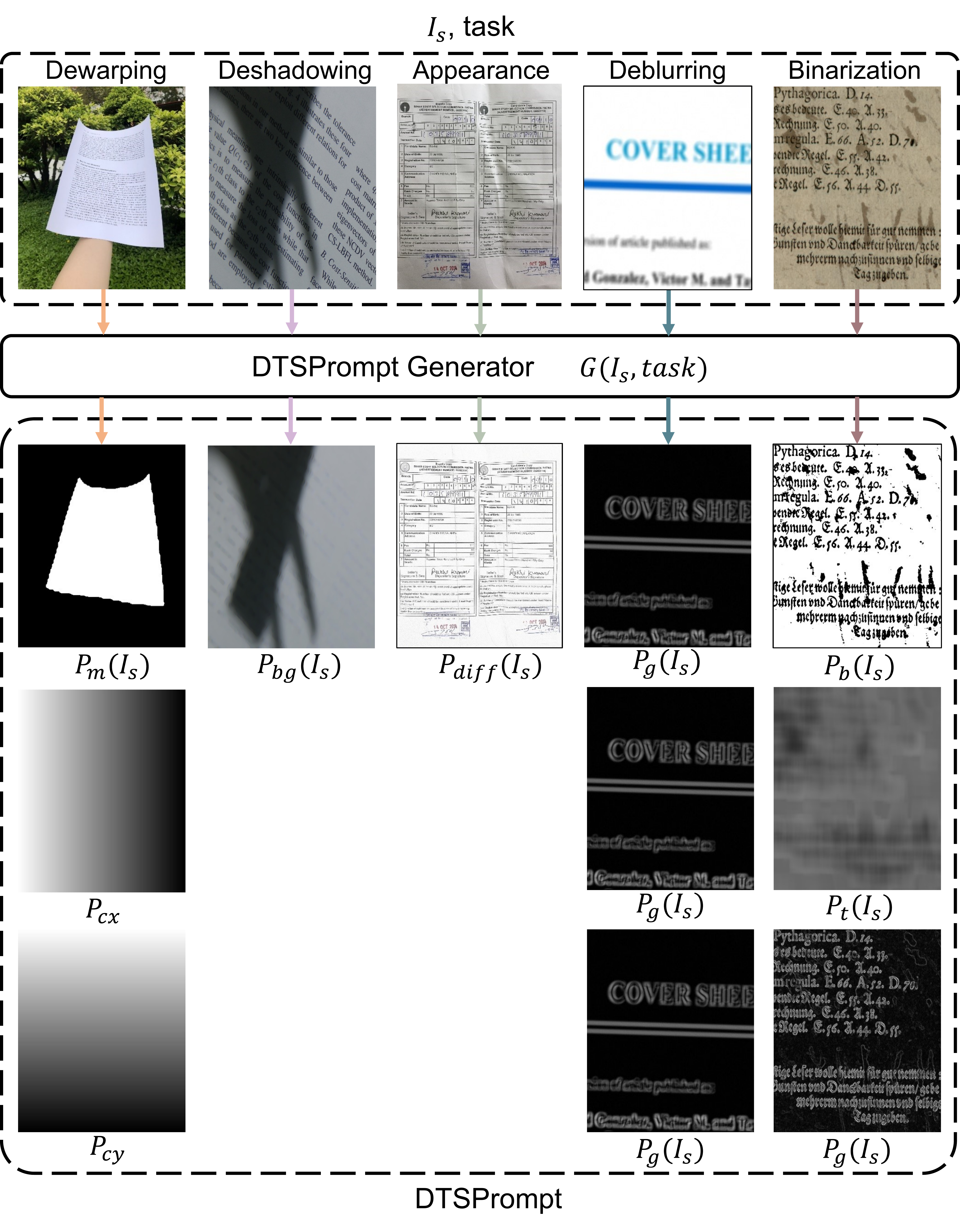}
    \centering
    \caption{The DTSPrompt for different tasks is composed of distinct prior features. Most of these prior features are extracted from the input image, making them dynamic. Zoom in for the best view.}
    \label{fig:dtsprompt}
\end{figure}

Prevailing visual generalist models~\cite{wang2023images,ma2023prores,tang2023unifying,liu2023unifying,wang2022ofa} determine prompts solely based on the task, and are independent of the input image:
\begin{equation}
\centering
  \label{common_prompt}
  { Prompt }=f(\text { task }).
\end{equation}
In contrast, our DTSPrompt dynamically adapts to the input image $I_{s}\in \mathbb{R}^{h \times w \times 3}$:
\begin{equation}
\centering
  \label{common_prompt}
  { DTSPrompt }=G\left(I_{s}, \text { task }\right) \in \mathbb{R}^{h \times w \times 3}.
\end{equation}
Here, $G$ represents our DTSPrompt generator (depicted in Fig.~\ref{fig:architecture}), which is responsible for extracting prior features from $I_s$ based on the specified task. In the following, we provide a detailed explanation of how the DTSPrompt for each task is constructed. Visual results of DTSPrompts for each task are presented in Fig.~\ref{fig:dtsprompt}.

\textbf{Dewarping}. Existing document dewarping methods~\cite{feng2022geometric,li2023foreground,jiang2022revisiting} often use text line masks or text block masks to assist the dewarping model, making it more attentive to the dewarping of regions with meaningful content. Document mask~\cite{zhang2022marior,feng2021doctr,li2023foreground,feng2022geometric} is commonly employed to enhance the model's understanding of document boundaries and reduce the learning difficulty by decoupling the margin removal and content rectification processes. Here, we choose the simplest document mask as our prior feature for the dewarping task. This mask, denoted as $P_m(I_s) \in \mathbb{R}^{h \times w}$, is obtained by directly using an existing document segmentation model proposed in~\cite{zhang2022marior} (noted that any other existing document segmentation models~\cite{feng2021doctr,feng2022geometric} can also be employed).

Furthermore, considering that predicting the backward map in dewarping task is inherently a problem related to coordinate positions, inspired by~\cite{liu2018intriguing,das2019dewarpnet}, we introduce the x-coordinate and y-coordinate as additional prior features, denoted as $P_{cx}\in \mathbb{R}^{h \times w}$ and $P_{cy}\in \mathbb{R}^{h \times w}$, respectively. They represent the coordinate values of the pixel at $(i, j)$, i.e., $P_{cx}(i, j)=i$ and $P_{cy}(i, j)=j$, which enable better perception of positional information. The DTSPrompt for dewarping task is obtained by concatenating these prior features along the channel dimension:
\begin{equation}
\centering
G\left(I_{s},``dewarp"\right)=\left[P_{m}\left(I_{s}\right), P_{c x}, P_{c y}\right].
\end{equation}

\textbf{Deshadowing}.  For the deshadowing task, the document background is commonly considered as a prior feature to enhance performance~\cite{lin2020bedsr,zhang2023document,liu2023shadow}. Here, we adopt the document background with shadows as our prior feature. To obtain this background from the document image, we initially employ dilation operations to eliminate the textual content within the document. Subsequently, we use a median filter to smooth out artifacts introduced due to incomplete removal. We represent this whole process as $P_{bg}(I_s)\in \mathbb{R}^{h \times w \times 3}$. The DTSPrompt for the deshadowing task is thus formulated as
\begin{equation}
\centering
G\left(I_{s},``deshadow"\right)=P_{bg}\left(I_{s}\right).
\end{equation}

\textbf{Appearance enhancement}. Current methods~\cite{zhang2023appearance,wang2022udoc,das2020intrinsic} commonly adopt background light, shadow map, or white-balance kernel as prior features to facilitate appearance enhancement based on the concept of intrinsic images. However, accurately obtaining these prior features is challenging and usually requires an extra model for training and predicting. For simplicity, we leverage the discrepancy between the original image and the document background $P_{bg}(I_s)$ as our prior feature for this task, which can serve as initial enhancement guidance for the model: 
\begin{equation}
\centering
 P_{diff}(I_s)=255-abs(I_s-P_{bg}(I_s)),
\end{equation}
\begin{equation}
\centering
G\left(I_{s},``appearance"\right)=P_{diff}\left(I_{s}\right).
\end{equation}

\textbf{Deblurring}. The gradient distribution is a prior feature widely used in traditional optimization-based deblurring methods~\cite{koh2021single}, which is typically employed as regularization information to constrain the solution space of the optimization function. 
Here, we use the gradient map of the input image $P_{g}\left(I_{s}\right) \in \mathbb{R}^{h \times w}$ as an additional input, aiming for the model to implicitly learn gradient prior information, rather than utilizing a gradient distribution prior to constrain the solution space of the output. The DTSPrompt for the deblurring task is expressed as
\begin{equation}
\centering
G\left(I_{s},``deblur"\right)=\left[P_{g}\left(I_{s}\right), P_{g}\left(I_{s}\right), P_{g}\left(I_{s}\right)\right].
\end{equation}

\textbf{Binarization}. Integrating prior features as supplementary for performance enhancement is widely adopted in binarization task~\cite{tensmeyer2017document, jia2018degraded, yang2024gdb}, and the effectiveness has been extensively demonstrated. For this task, we employ the Sauvola binarization algorithm~\cite{sauvola2000adaptive} to yield the initial binarization outcome and threshold map as our prior features, which is denoted as $P_{b}(I_s)\in \mathbb{R}^{h \times w}$ and $P_{t}(I_s)\in \mathbb{R}^{h \times w}$, respectively. Furthermore, we also incorporate gradient information as an additional prior feature for this task. The DTSPrompt for binarization task is formulated as
\begin{equation}
\centering
G\left(I_{s},``binarize"\right)=\left[P_{b}\left(I_{s}\right), P_{t}\left(I_{s}\right), P_{g}\left(I_{s}\right)\right].
\end{equation}

\subsection{Prompt fusion and restoration network}
\label{sec:prompt}

Exploring how to integrate the acquired prompt information into the network holds significant merit. A well-conceived fusion method has the potential to substantially enhance overall performance. However, the primary focus of this paper lies in evaluating the efficacy and potentials associated with DTSPrompt, rather than delving into the intricacies of designing a complex network structure. In line with this objective, we adopt a straightforward fusion approach to seamlessly incorporate DTSPrompt into the restoration network. Specifically, we opt for concatenating DTSPrompt and $I_s$ along the channel dimension to construct a new input $\in \mathbb{R}^{h \times w \times 6}$ for the restoration network.

Due to the simplicity of DTSPrompt, we have the flexibility to choose from various restoration networks. In this case, we opt for the off-the-shelf Restormer~\cite{zamir2022restormer} without modifications to form our DocRes. With such a restoration network, DocRes can support inputs of up to $1600\times1600$ and adapt to inputs with variable resolutions. It's noteworthy that other restoration networks can be employed interchangeably since DTSPrompt does not require specific modules within the network.

\section{Experiments}
\subsection{Datasets}
\label{set:dataset}

\textbf{Dewarping}. We adopt the Doc3D~\cite{das2019dewarpnet} dataset and the DIR300 benchmark~\cite{feng2022geometric} for the training and testing, respectively. Doc3D is a synthetic dataset comprising 100K samples, which includes geometrically distorted document images and corresponding backward maps. DIR300 is a real-world benchmark with 300 geometrically distorted images and corresponding flat ground-truths.

\textbf{Deshadowing}. The training set for this task consists of 14,200 synthetic images from FSDSRD~\cite{matsuo2022document} and 4,371 real images from the training set of RDD~\cite{zhang2023document}. We use Jung's dataset~\cite{jung2018water} (87 images), Kligler's dataset~\cite{kligler2018document} (300 images) and OSR~\cite{wang2020local} (237 images) to form our testing set.


\textbf{Appearance enhancement}. The training set for this task contains 90K synthetic images from the Doc3DShade~\cite{das2020intrinsic} dataset and 450 real-world images from the RealDAE~\cite{zhang2023appearance} training set. 150 images from the testing set of RealDAE and 130 images from DocUNet~\cite{ma2018docunet} are used for evaluation. As introduced in~\cite{feng2021doctr,das2019dewarpnet}, the degraded images in DocUNet should be aligned to the flat ground truths before the evaluation of this task. Following~\cite{zhang2023appearance}, we achieve alignment by using the document alignment model~\cite{zhang2023docaligner} rather than some dewarping models~\cite{zhang2022marior,feng2021doctr,ma2018docunet}, which can result in better alignment and thus provide a more accurate evaluation. We denote the aligned dataset as DocUNet*.

\textbf{Deblurring}. The Text Deblur Dataset (TDD)~\cite{hradivs2015convolutional} consists of 66K training samples, from which we randomly select 40K samples to train our model. The 1.6K testing samples from TDD form the testing set of this task.

\textbf{Binarization}. We use DIBCO'18~\cite{2018ICFHR} as our testing set. Following~\cite{yang2023docdiff,yang2024gdb}, the remaining years of (H)-DIBCO datasets~\cite{pratikakis2017,pratikakis2016icfhr2016,ntirogiannis2014icfhr2014,pratikakis2013icdar,pratikakis2012icfhr,pra2011icdar,pratikakis2010h,gatos2009icdar,2019ICDAR} are used as the training data, and images from Noisy Office dataset~\cite{Francisco2007Behaviour}, Synchromedia Multispectral dataset~\cite{hedjam2015icdar}, Persian Heritage Image Binarization dataset~\cite{nafchi2013efficient} and Bickley Diary dataset~\cite{deng2010binarizationshop} are also used for training.

\begin{table*}[t]
\caption{Quantitative comparison results between our all-in-one generalist DocRes model and existing task-specific state-of-the-art models on 5 tasks. From top to bottom, the tasks include dewarping, deshadowing, appearance enhancement, deblurring, and binarization. \textbf{Best} results are shown in bold.}
\label{tab:sota}
\begin{adjustbox}{width=1\textwidth}
\renewcommand{\arraystretch}{1}
\begin{tabular}{c|c|ccccccccc|c}
\Xhline{1px}
 Metrics & Datasets & \makecell{DocGeo~\cite{feng2022geometric}\\ \textit{ECCV'22}} & \makecell{Li et al.~\cite{li2023foreground}\\ \textit{ICCV'23}} & \makecell{BGSNet~\cite{zhang2023document}\\\textit{CVPR'23}} & \makecell{DocShadow~\cite{li2023high}\\\textit{ICCV'23}} & \makecell{UDoc-GAN~\cite{wang2022udoc}\\\textit{MM'22}} & \makecell{GCDRNet~\cite{zhang2023appearance}\\\textit{TAI'23}} & \makecell{GDB~\cite{yang2024gdb}\\\textit{PR'24}} & \makecell{DE-GAN~\cite{souibgui2020gan}\\\textit{TPAMI'20}} & \makecell{DocDiff~\cite{yang2023docdiff}\\\textit{MM'23}} & \makecell{DocRes\\(\textbf{ours})} \\ \hline
\makecell{MSSSIM$\uparrow$\\AD$\downarrow$\\LD$\downarrow$} & DIR300~\cite{feng2022geometric} & \makecell{\textbf{0.6380}\\0.242\\\textbf{6.40}} & \makecell{0.6070\\0.244\\7.68} & - & - & - & - & - & - & - & \makecell{0.6264\\\textbf{0.241}\\6.83} \\ \hline
\multirow{3}{*}[-3ex]{\makecell{SSIM$\uparrow$\\PSNR$\uparrow$}} & Kligler et al.~\cite{kligler2018document} & - & - & \makecell{\textbf{0.9480}\\\textbf{29.17}} & \makecell{0.9088\\25.12} & - & - & - & - & - & \makecell{0.9005\\27.14} \\ \cline{2-12} 
 & Jung et al.~\cite{jung2018water} & - & - & \makecell{0.9040\\17.34} & \makecell{0.9005\\21.05} & - & - & - & - & - & \makecell{\textbf{0.9089}\\\textbf{23.02}}  \\ \cline{2-12} 
 & OSR~\cite{wang2020local} & - & - & \makecell{\textbf{0.9388}\\\textbf{22.64}} & \makecell{0.9023\\18.25} & - & - & - & - & - & \makecell{0.9370\\21.64} \\ \hline
\multirow{2}{*}[-1.8ex]{\makecell{SSIM$\uparrow$\\PSNR$\uparrow$}} & DocUNet*~\cite{ma2018docunet} & - & - & - & - & \makecell{0.6833\\14.29} & \makecell{\textbf{0.7658}\\17.09} & - & - & - & \makecell{0.7598\\\textbf{17.60}} \\ \cline{2-12} 
 & RealDAE~\cite{zhang2023appearance} & - & - & - & - & \makecell{0.7558\\16.43} & \makecell{\textbf{0.9423}\\24.42} & - & - & - & \makecell{0.9219\\\textbf{24.65}} \\ \hline
\makecell{SSIM$\uparrow$\\PSNR$\uparrow$} & TDD~\cite{hradivs2015convolutional} & - & - & - &  - & - & - & - & \makecell{0.9226\\22.24} & \makecell{0.9559\\24.00} & \makecell{\textbf{0.9723}\\\textbf{27.35}} \\ \hline
\makecell{FM$\uparrow$\\pFM$\uparrow$\\PSNR$\uparrow$} & DIBCO'18~\cite{2018ICFHR} & - & - & - & - & - & - & \makecell{\textbf{91.09}\\\textbf{94.57}\\\textbf{19.92}} & \makecell{77.59\\85.74\\16.16}& \makecell{88.11\\90.43\\17.92} & \makecell{89.82\\94.33\\19.35} \\ \cline{2-12} \Xhline{1px}
\end{tabular}
\end{adjustbox}
\end{table*}

\begin{table}[]\footnotesize
\centering
\caption{Quantitative results of the ablation experiments. \textbf{Best} results are shown in bold.}
\begin{adjustbox}{width=0.48\textwidth}
\renewcommand{\arraystretch}{1}
\renewcommand{\multirowsetup}{\centering}
\label{tab:ablation}
\begin{tabular}{c|c|c|ccc}
\Xhline{1px}
\multirow{2}{*}[-3ex]{Metrics} & \multirow{2}{*}[-3ex]{Datasets} & \makecell{Task\\specific} & \multicolumn{3}{c}{Unified} \\ \cline{3-6} 
 &  & Baseline & Baseline & \makecell{Baseline\\+Fix prompt} & \makecell{Baseline\\+DTSPrompt\\(DocRes)} \\ \hline
\makecell{MSSSIM$\uparrow$\\AD$\downarrow$\\LD$\downarrow$} & DIR300~\cite{feng2022geometric} & \makecell{0.6185\\0.263\\7.18} & \makecell{0.5943\\0.333\\8.26} & \makecell{0.6030 \\0.302\\7.77} & \textbf{\makecell{0.6264\\0.241\\6.83}} \\ \hline
\multirow{3}{*}[-3ex]{\makecell{SSIM$\uparrow$\\PSNR$\uparrow$}} & Kilgler et al.~\cite{kligler2018document} & \makecell{0.8879\\\textbf{27.41}} & \makecell{0.8238\\11.95} & \makecell{0.8977\\27.08} & \makecell{\textbf{0.9005}\\27.14} \\ \cline{2-6} 
 & Jung et al.~\cite{jung2018water} & \makecell{0.9025\\\textbf{23.18}} & \makecell{0.8573\\18.01} & \makecell{0.9031\\22.42} & \makecell{\textbf{0.9089}\\23.02} \\ \cline{2-6} 
 & OSR~\cite{wang2020local} & \makecell{0.9259\\20.12 } & \makecell{0.8634\\12.35} & \makecell{0.9285\\19.47} & \textbf{\makecell{0.9370\\21.64}} \\ \hline
\multirow{2}{*}[-2ex]{\makecell{SSIM$\uparrow$\\PSNR$\uparrow$}} & DocUNet*~\cite{ma2018docunet} & \makecell{0.7621\\17.43} & \makecell{0.7620\\17.30} & \makecell{\textbf{0.7635}\\17.28} & \makecell{0.7598\\\textbf{17.60}} \\ \cline{2-6} 
 & RealDAE~\cite{zhang2023appearance} & \makecell{0.9204\\24.21} & \makecell{0.9050\\21.75} & \makecell{0.9200\\24.32} & \makecell{\textbf{0.9219}\\\textbf{24.65}} \\ \hline
\makecell{SSIM$\uparrow$\\PSNR$\uparrow$} & TDD~\cite{hradivs2015convolutional} & \textbf{\makecell{0.9811\\28.95}} & \makecell{0.9639\\25.85} & \makecell{0.9690\\26.64} & \makecell{0.9723\\27.35} \\ \hline
\makecell{FM$\uparrow$\\pFM$\uparrow$\\PSNR$\uparrow$} & DIBCO'18~\cite{2018ICFHR} & \makecell{76.57\\79.51\\14.64} & \makecell{74.33\\76.15\\14.47} & \makecell{77.15\\81.18\\15.28} & \textbf{\makecell{89.82\\94.33\\19.35}} \\ \Xhline{1px}
\end{tabular}
\end{adjustbox}
\end{table}

\subsection{Evaluation metrics}
Deshadowing, appearance enhancement, and deblurring tasks adopt the commonly used PSNR and SSIM as evaluation metrics. The evaluation of dewarping incorporates multi-scale structural similarity (MS-SSIM)~\cite{wang2003multiscale}, local distortion (LD)~\cite{you2017multiview} and align distortion (AD)~\cite{ma2022learning}. MS-SSIM builds upon the traditional SSIM by considering multiple scales. LD evaluates dewarping performance by utilizing the offset between the dewarped result and the flat ground truth. AD, an enhancement of LD, refines the evaluation by excluding offset noise in low-textured regions and mitigating the impact of global transformations. the parameters for MS-SSIM, LD, and AD align with those established in prior works~\cite{ma2018docunet,das2019dewarpnet,feng2022geometric,li2023foreground}. For the binarization task, we employ PSNR, F-measure (FM), and pseudo F-measure (pFM) as our evaluation metrics.

\subsection{Implementation details}
We train our model on 8 NVIDIA A6000 GPUs for 100,000 steps with a global batch size of 80. AdamW with a weight decay of $5 \times 10^{-4}$ is adopted. We use the cosine learning rate scheduler with $2 \times 10^{-4}$ as the maximum learning rate. 

Before commencing such unified training, the model undergoes a pre-training phase exclusively on the dewarping task for 50,000 steps to initialize the model. This is for a more stable training purpose, as dewraping significantly differs from other tasks: while it involves coordinates regression, other tasks entail regression of image content. 

During the unified training process, the sampling weight for dewarping, deshadowing, appearance enhancement, deblurring, and binarization are all set to 0.2. Apart from the binarization task, which employs the standard cross-entropy loss for its output, all other tasks are supervised using the L1 loss. Images of deshadowing, appearance enhancement, deblurring, and binarization tasks are randomly cropped as patches with a size of $256 \times 256$, while images of dewarping task are resized to $256 \times 256$ during training.

\subsection{Results}

\textbf{Comparisons with SOTA task-specific models}. We conduct a comprehensive comparison between our proposed DocRes and existing meticulously designed task-specific methods. Specifically, for the dewarping task, we benchmark DocRes against the current state-of-the-art method, DocGeo~\cite{feng2022geometric}, and the recent model by Li et al.~\cite{li2023foreground}. In the deshadowing domain, we compare DocRes with the latest SOTA models, namely BGSNet~\cite{zhang2023document} (utilizing the model trained on the RDD dataset provided by the authors) and DocShadow~\cite{li2023high} (based on the model trained on the SD7K dataset provided by the authors). For appearance enhancement, we assess DocRes against UDoc-GAN~\cite{wang2022udoc} and GCDRNet~\cite{zhang2023appearance}. The binarization task involves a comparison with the current SOTA method, GDB~\cite{yang2024gdb}. Additionally, we contrast our approach with DocDiff~\cite{yang2023docdiff} and DE-GAN~\cite{souibgui2020gan}, both of which aim to unify multiple tasks within a single network structure but still necessitate separate training for each task, resulting in the need for multiple models.

\begin{figure*}[t]
    \includegraphics[width=6.2in]{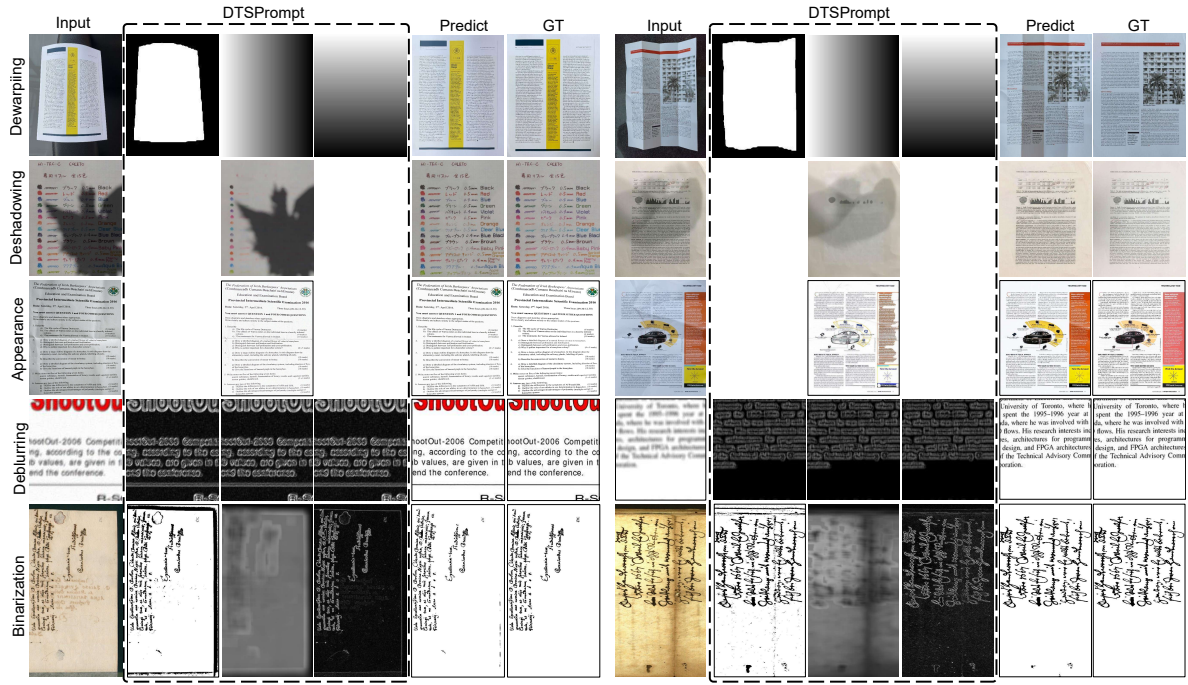}
    \centering
    \caption{Visualizations showcasing inputs, DTSPrompt, and restoration results from DocRes, and ground truths across various tasks, including dewarping, deshadowing, appearance enhancement, deblurring, and binarization. Zoom in for the best view.}
    \label{fig:visualize}
\end{figure*}

Results across multiple benchmark datasets for the five tasks are presented in Table~\ref{tab:sota}. It can be seen that DocRes not only competes with existing task-specific SOTA models but also surpasses them in several instances. DocRes achieves new records in certain metrics for benchmark datasets related to dewarping, deshadowing, deblurring, and appearance enhancement tasks. Even for binarization tasks, where the dedicated SOTA model GDB still holds the top position, DocRes exhibits performance closely trailing behind it. In contrast to existing unified-structure methods like DE-GAN and DocDiff, which still require separate training for each task, DocRes demonstrates significant advantages. Visualized results on these benchmarks from DocRes are showcased in Fig.~\ref{fig:visualize}.

\begin{figure}[t]
    \includegraphics[width=2.8in]{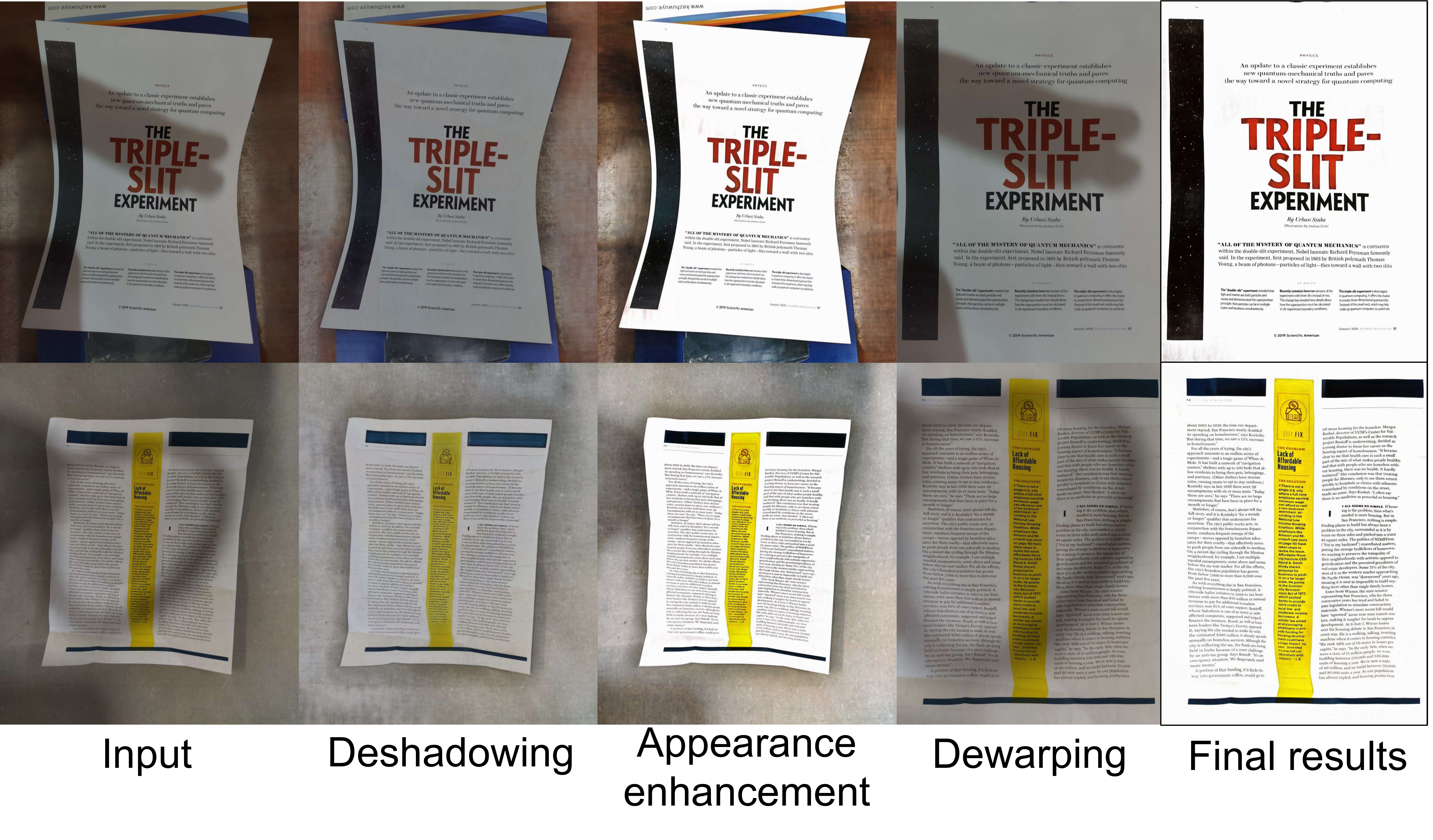}
    \centering
    \caption{For the same input, we employ different DTSPrompts to validate the controllability of DocRes. The results in the last column indicate that using DocRes enables the complete restoration process for photographed documents, achieving the desired final results for users. Zoom in for the best view.}
    \label{fig:control}
\end{figure}

While there remains room for improvement compared to some of these well-designed specialized models, it's important to underscore that our primary objective in this paper is not to achieve SOTA performance on every task. Instead, the focus is on evaluating the efficacy and potential of the unified DocRes approach. Further research, for example, can explore more sophisticated prompt fusion mechanisms to achieve SOTA performance in each task.

\textbf{Ablation studies}. In this subsection, we conduct ablation studies to evaluate the effectiveness of our DTSPrompt. Restormer~\cite{zamir2022restormer} is treated as our baseline model. We first individually train it on each task, obtaining task-specific results shown in the third column of Table~\ref{tab:ablation}. As a state-of-the-art image restoration network, Restormer demonstrates proficiency across various tasks when trained in isolation. However, when training it in a unified model setting (the fourth column of Table~\ref{tab:ablation}), we see a significant performance decline across almost all benchmarks. This decline is attributed to the similarity in input across tasks, coupled with distinct output requirements, leading to confusion in the model's learning process.

Additionally, we explore the effectiveness of fixed prompts. A fixed prompt for each task is a $h\times w\times 3$ matrix, with constant values determined solely by the task and remaining constant regardless of the input image. For example, the fixed prompt for the dewarping task is a $h\times w\times 3$ matrix filled with zeros, while for the deshadowing task, it is filled with ones. We adopt the same fusion approach, where the concatenated result of the fixed prompt and the input image are fed into the restoration network.

Such a simple fixed prompt approach achieves competitive performance compared to task-specific settings in appearance enhancement and deshadowing tasks. There is even a noticeable improvement in the binarization task, which could be attributed to multi-task learning helping mitigate the generalization issue caused by the scarcity of training data for the binarization task. However, for tasks like dewarping and deblurring, the fixed prompt method still experiences significant performance declines, highlighting the challenges of creating an excellent generalist model.

Our DTSPrompt excels in dewarping, deshadowing, deblurring, and binarization tasks compared to the fixed prompt approach. Particularly, there are significant improvements in the dewarping and binarization tasks. From an overall perspective across all tasks, DTSPrompt achieves superior results compared to the task-specific setting, by using only a single model without the need for additional training parameters or structural modifications.

\textbf{Control ability}. In this subsection, we explore the control ability of DTSPrompt. Our focus is on observing how well DocRes can perform the correct restoration task when different DTSPrompts are employed for the same input image. Specifically, we consider three tasks related to the photographed scene: dewarping, deshadowing, and appearance enhancement. The visualization results in the first four columns of Fig.~\ref{fig:control} show that DocRes can accurately perform the corresponding tasks. In addition, we also demonstrate in the last column that a single DocRes model can complete the entire enhancement process of photographed document images and obtain the final results desired by the user.

\begin{figure}[t]
    \includegraphics[width=2.5in,height=2.2in]{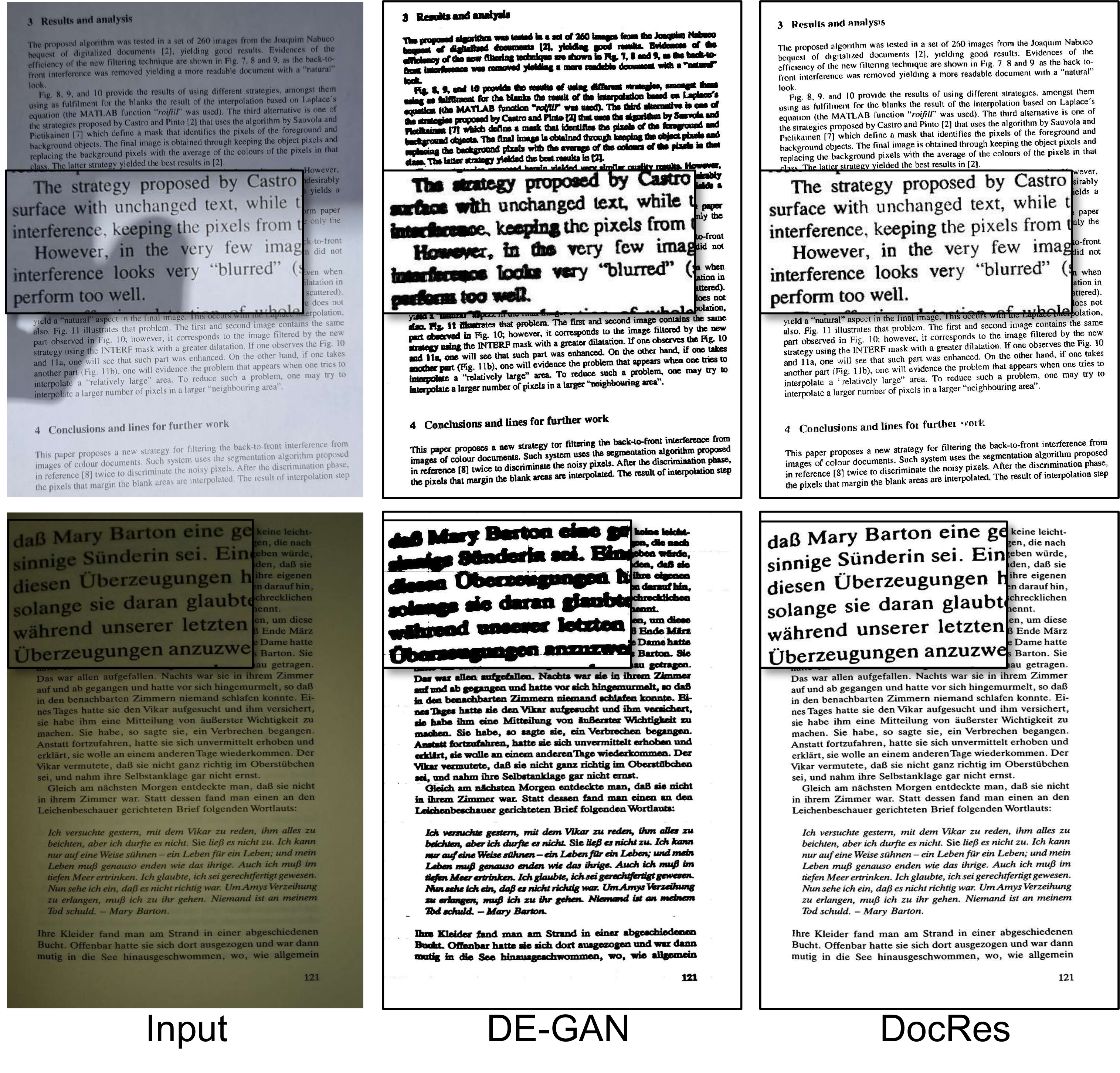}
    \centering
    \caption{Visualization results when applying models (DE-GAN~\cite{souibgui2020gan} and our DocRes) to perform binarization on photographed documents, which were merely trained on scanned ancient document binarization data. Zoom in for the best view.}
    \label{fig:generalization1}
\end{figure}

\textbf{Generalization}. An essential trait of a generalist model lies in its ability to harness synergies among multi-task data to improve overall generalization. In this context, we delve into the generalization capability of DocRes through visualizations. As outlined in Section~\ref{set:dataset}, the training data for the binarization task in DocRes primarily consists of scanned documents, particularly ancient ones, presenting a considerable gap from the photographed modern documents~\cite{bernardino2023quality}. Here, we aim to apply DocRes to binarize photographed documents, introducing unseen noises like blurriness, shadows, low-light conditions, and reflections that are not present in the binarization training set. As shown in Fig.~\ref{fig:generalization1}, DocRes consistently demonstrates excellent performance on such out-of-domain data. In contrast, DE-GAN's~\cite{souibgui2020gan} performance significantly deteriorates in the presence of shadow and low-light interference.

\begin{figure}[t]
    \includegraphics[width=3.2in]{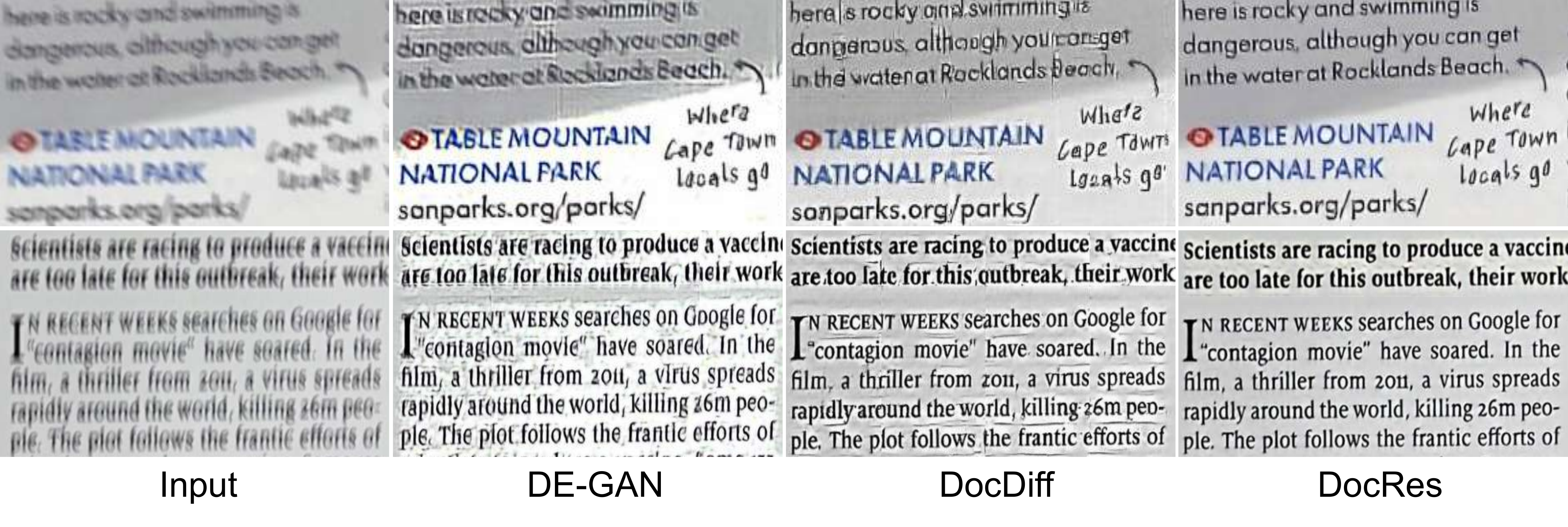}
    \centering
    \caption{Visualization results when applying models (DE-GAN~\cite{souibgui2020gan}, DocDiff~\cite{yang2023docdiff} and our DocRes) to perform the deblurring task on photographed documents, which were merely trained on clean document blurring data. Zoom in for the best view.}
    \label{fig:generalization2}
\end{figure}

Moreover, we extend the evaluation to the deblurring task for photographed document images. Notably, the training data for the deblurring task in DocRes consists exclusively of clean document images. As illustrated in Fig.~\ref{fig:generalization2}, DocRes exhibits superior deblurring performance on photographed document images compared to DE-GAN~\cite{souibgui2020gan} and DocDiff~\cite{yang2023docdiff}, both of which were also trained exclusively on the TDD dataset for the deblurring task.

We attribute DocRes's robust out-of-domain generalization capability to its learning of patterns associated with photographed noise through tasks like dewarping, deshadowing, and appearance enhancement.

\section{Discussions and conclusions}

This paper presents DocRes, a generalist model designed for unifying document image restoration tasks, including dewarping, deshadowing, appearance enhancement, deblurring, and binarization. The key innovation of DocRes is the incorporation of Dynamic Task-Specific Prompt (DTSPrompt), which leverages prior features to construct visual prompts, acting not only as a guiding cue for specific tasks but also supplying additional information to enhance restoration performance. It can be seamlessly applied to existing restoration networks, resulting in a generalist model that can accommodate input with high or variable resolutions. With the support of DTSPrompt, DocRes achieves performance levels matching or surpassing SOTA task-specific models, without the need for extra training parameters or complex architectural designs. We also illustrate DocRes's controllability in performing different tasks when presented with the same input image and its capacity to generalize to out-of-domain data.

Notably, the DTSPrompt is not confined to the specific tasks explored in this paper. It can be potentially extended to incorporate more diverse prior features~\cite{huang2010detecting,luo2010jpeg,yoo2018image,qu2023towards}, such as DCT coefficients, SIFT, JPEG noise, and resampling artifacts, to accommodate a broader range of image restoration tasks. Additionally, it is worth investigating prompt fusion mechanisms for better integrating the DTSPrompt. In summary, this paper successfully attempts to develop a unified multi-task model for document image restoration, inspiring future research of generalist or foundation models for pixel-level image processing tasks.  

\section*{Acknowledgement}
This research is supported in part by the National Natural Science Foundation of China (Grant No.: 62441604, 61936003), the National Key Research and Development Program of China  (2022YFC3301703)

{
    \small
    \bibliographystyle{ieeenat_fullname}
    \bibliography{main}
}

\clearpage
\setcounter{page}{1}
\setcounter{figure}{0}
\setcounter{table}{0}
\setcounter{section}{0}
\maketitlesupplementary

\section{Efficiency}
As shown in Table~\ref{tab:model_size}, we compared the number of parameters and computational complexities of our method with those of other methods. It can be observed that even when compared to certain task-specific models, our method maintains an advantage in both the number of parameters and computational complexity. This advantage is particularly significant when considering that the number of parameters of these task-specific models would multiply several times over if they were to support multitasks. In contrast, our method does not require additional parameter increments.

\begin{table}[h]
\centering
\caption{Model size and computational complexities for each method.}
\renewcommand{\arraystretch}{1}
\resizebox{0.47\textwidth}{!}{
\begin{tabular}{c|ccccc}
\hline
Methods & Li et al.~\cite{li2023foreground} & DocShadow~\cite{li2023high} & UDoc-GAN~\cite{wang2022udoc} & DE-GAN~\cite{souibgui2020gan} & DocRes (ours)\\ \hline
Params (M) $\downarrow$ & 63.5 & 29.3 & 19.6 & 30 & \textbf{15.2} \\
GFLOPs $\downarrow$ & 262.4 & 9.7 & \textbf{2.0} & 109.0 & 183.0 \\ \hline
\end{tabular}}
\label{tab:model_size}
\end{table}

\section{Additional ablation study}
To quantify the individual contributions of task synergies and DTSPrompt, we conducted an additional ablation experiment and presented the results in Table \ref{tab:synergy}. From the table, it can be observed that both DTSPrompt and multitask synergy lead to improvements. The improvement brought by multitask synergy may be attributed to the relatively small amount of binarization training data. The incorporation of other task data aids the model in avoiding overfitting to the limited binarization training data.

\begin{table}[h]
\centering
\caption{For binarization, we gradually add DTSPrompt and multi-task to provide further insights of these factors.}
\renewcommand{\arraystretch}{1}
\resizebox{0.38\textwidth}{!}{
\begin{tabular}{cc|c}
\hline
DTSPrompt & Multi-task & DIBCO'18 (FM$\uparrow$/pFM$\uparrow$/PSNR$\uparrow$) \\ \hline
\XSolidBrush & \XSolidBrush & 76.5 / 79.5 / 14.4 \\
\Checkmark & \XSolidBrush & 87.1 / 90.6 / 18.6 \\
\Checkmark & \Checkmark & \textbf{89.8} / \textbf{94.3} / \textbf{19.3} \\ \hline
\end{tabular}}
\label{tab:synergy}
\end{table}

\section{Comparison with more SOTA}
We provide quantitative comparisons with more state-of-the-art (SOTA) methods for dewarping task in Table~\ref{tab:more_sota}, which demonstrates the superior performance of DocRes when compared with some of these task-specific methods.

\begin{table}[]
\centering
\caption{Quantitative comparison with existing SOTA task-specific document dewarping methods on the DIR300~\cite{feng2022geometric} benchmark.}
\resizebox{0.38\textwidth}{!}{
\begin{tabular}{c|c|ccc}
\hline
Methods & Venue & MS-SSIM$\uparrow$ & LD$\downarrow$ & AD$\downarrow$ \\ \hline
DewarpNet~\cite{feng2021doctr} & ICCV'19 & 0.492 & 13.94 & 0.254 \\
DDCP~\cite{xie2021document} & ICDAR'21 & 0.552 &	10.95 & 0.331 \\
DocTr~\cite{feng2021doctr} & MM'21 & 0.616 & 7.21 & 0.254 \\
DocGeoNet~\cite{feng2022geometric} & ECCV'22 & \underline{0.638} & \underline{6.40} & 0.242 \\
PaperEdge~\cite{ma2022learning} & SIGGRAPH'22 & 0.583 & 8.00 & 0.255 \\
Li et al.~\cite{li2023foreground} & ICCV'23 & 0.607 & 7.68 & 0.244 \\
LA-DocFlatten~\cite{li2023layout} & ACM TOG'23 & \textbf{0.651} & \textbf{5.70} & \textbf{0.195} \\ \hline
DocRes (ours) & - & 0.626 & 6.83 & \underline{0.241} \\ \hline
\end{tabular}}
\label{tab:more_sota}
\end{table}

\section{More visualized results}
\begin{figure}[t]
    \includegraphics[width=3.2in]{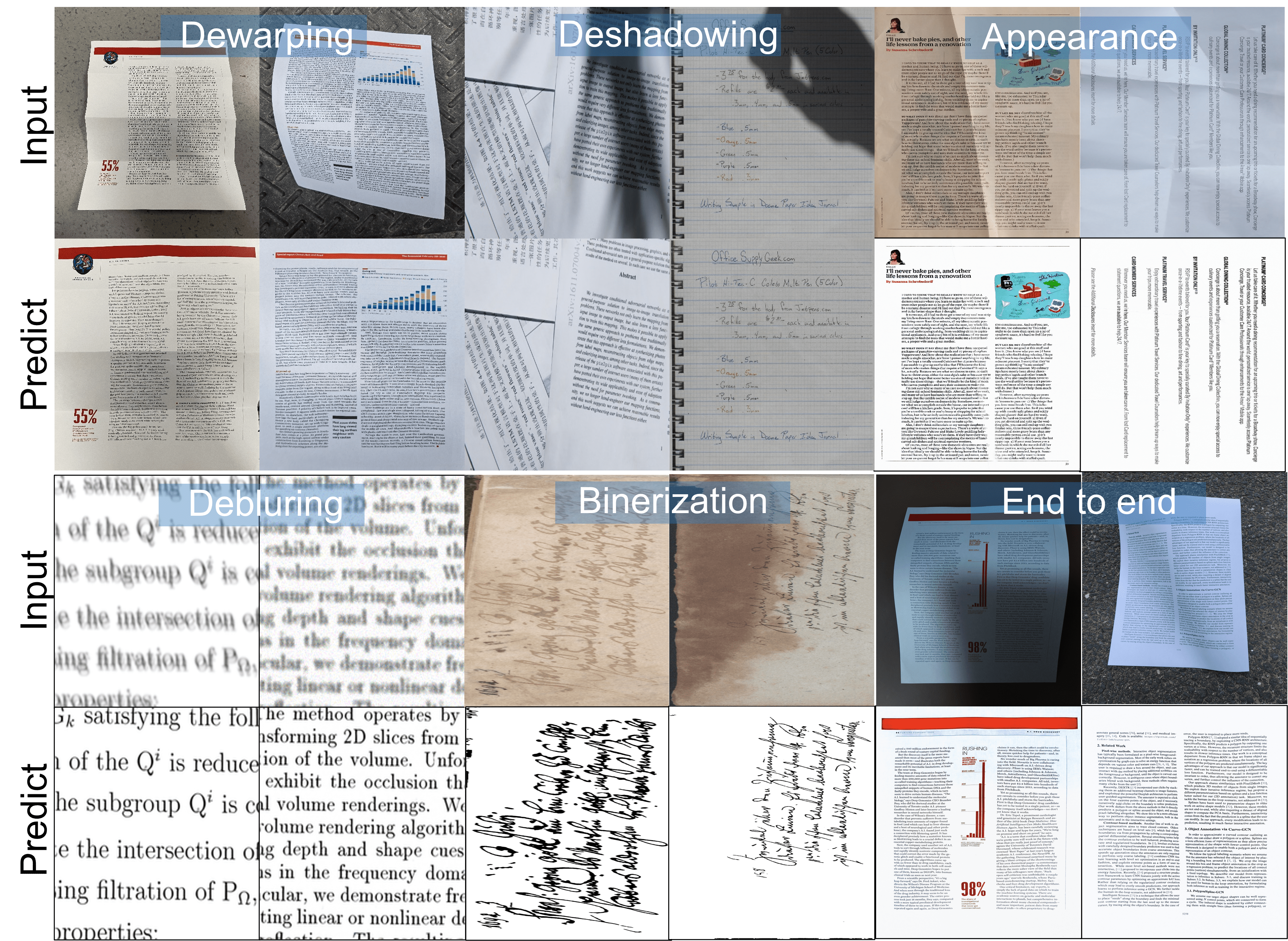}
    \centering
    \caption{More visualized results from DocRes. Zoom in for best view.}
    \label{fig:more_results}
\end{figure}

\begin{figure}[t]
    \includegraphics[width=3.2in]{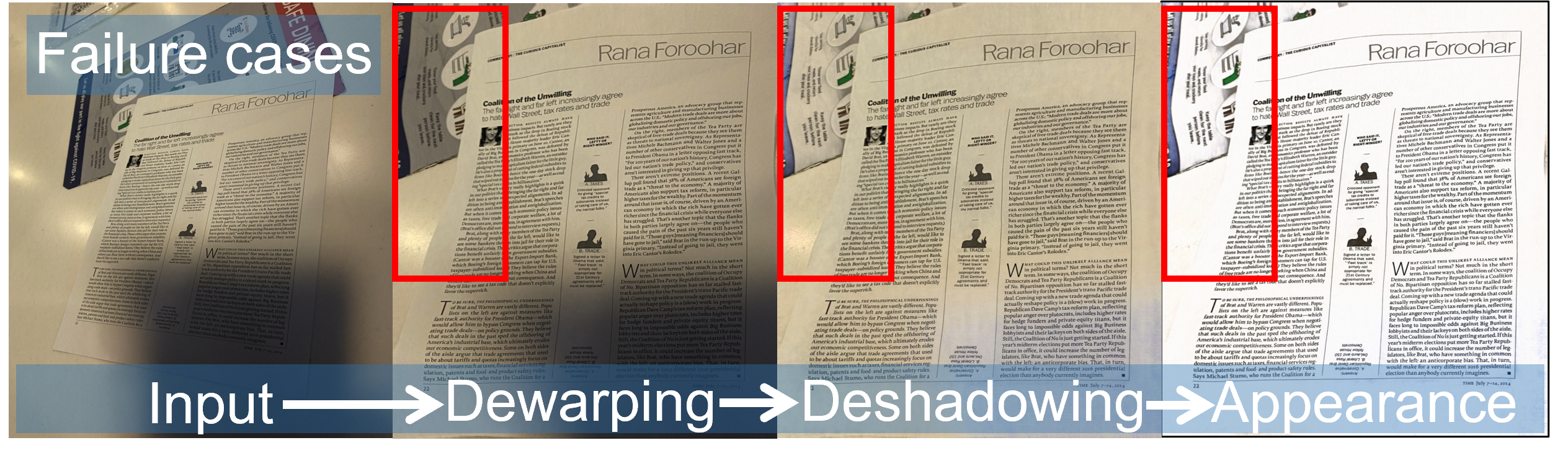}
    \centering
    \caption{Failure case from DocRes when applying it for end-to-end camera-captured document image enhancement. Zoom in for best view.}
    \label{fig:failure}
\end{figure}

In Fig.~\ref{fig:more_results}, we present additional visualization results of DocRes across multiple tasks. Additionally, in Fig.~\ref{fig:failure}, we illustrate the potential issue of error accumulation when applying DocRes for end-to-end camera-captured document image enhancement. Due to the iterative nature of DocRes for end-to-end tasks, where each forward pass utilizes the output of the previous pass as input, errors in one task can accumulate and affect the final result. Exploring methods that can accomplish multiple document restoration tasks in a single forward pass would be a meaningful avenue for future research.

\section{Further discussions about DTSPrompt}
In fact, from the ablation experiments in Table 2 of the main text, it is evident that compared to fixed prompts or task-specific models, DTPrompt does not demonstrate significant advantages and may even perform worse in certain tasks, such as shadow removal and deblurring tasks. This is mainly because document image restoration tasks like deblurring lack recognized prior features that definitively enhance performance. In such tasks, DTPrompt primarily serves as a discriminative cue rather than significantly boosting model performance. As for shadow removal tasks, while shadow maps have been widely proven to be useful, our extraction operations are very simple, involving basic image processing techniques rather than the conventional approach~\cite{zhang2023appearance,zhang2023document} of using deep models for prediction.

Based on these observations and analyses, future improvements for DTSPrompt could focus on two directions: \textbf{1.} Exploring more effective prior features for specific tasks: This entails delving deeper into identifying prior features that are more conducive to enhancing performance for particular tasks, such as deblurring. \textbf{2.} Employing a trainable prediction module for the DTSPrompt generator: This would enhance the prior feature extraction capabilities. Importantly, there's still no necessity to designate a separate DTSPrompt generator for each task. Instead, a single DTSPrompt generator with shared parameters could simultaneously output multiple prior features. During input to the restoration network, different prior features can be chosen for task guidance based on the specific task at hand.

\end{document}